\documentclass{article}
\usepackage{spconf,amsmath,graphicx}
\usepackage{times}
\usepackage{url}
\usepackage{latexsym}
\usepackage{xcolor}
\usepackage{color, colortbl}
\usepackage{multirow}
\usepackage{enumitem}

\usepackage{arabtex}	
\usepackage{graphicx}
\usepackage[export]{adjustbox}
\usepackage{utf8}

\definecolor{Gray}{gray}{0.95}

\title{The MGB-2 Challenge:\\ Arabic Multi-Dialect Broadcast Media Recognition}
\name{{\em Ahmed Ali$^{1,2}$, Peter Bell$^2$, James Glass$^3$, Yacine Messaoui$^{4}$, Hamdy Mubarak$^{1}$,  Steve Renals$^2$, Yifan Zhang$^{1}$\thanks{Authors listed alphabetically.}}}
\address{$^1$Qatar Computing Research Institute, HBKU, Doha, Qatar \\
  $^2$Centre for Speech Technology Research, University of Edinburgh, UK \\ 
  $^3$ MIT Computer Science and Artificial Intelligence Laboratory (CSAIL), Cambridge, MA, USA\\
  $^4$Aljazeera Media Network, Doha, Qatar \\
{\footnotesize  mgb-admin@inf.ed.ac.uk www.mgb-challenge.org}
  }
\begin{document}
%
\maketitle
\begin{abstract}
This paper describes the Arabic Multi-Genre Broadcast (MGB-2) Challenge for SLT-2016. Unlike last year's English MGB Challenge, which focused on recognition of diverse TV genres, this year, the challenge has an emphasis on handling the diversity in dialect in Arabic speech.  Audio data comes from 19 distinct programmes from the Aljazeera Arabic TV channel between March 2005 and December 2015. Programmes are split into three groups: conversations, interviews, and reports.  A total of 1,200 hours have been released with lightly supervised transcriptions for the acoustic modelling. For language modelling, we made available over 110M words crawled from Aljazeera Arabic website Aljazeera.net for a 10 year duration 2000-2011. Two lexicons have been provided, one phoneme based and one grapheme based. Finally, two tasks were proposed for this year's challenge: standard speech transcription, and word alignment. This paper describes the task data and evaluation process used in the MGB challenge, and summarises the results obtained.
\end{abstract}
\begin{keywords}
Speech recognition, broadcast speech, transcription, multi-genre, alignment
\end{keywords}
\section{Introduction}
\label{sec:intro}

The second round of the Multi-Genre Broadcast MGB~\cite{bell2015mgb} challenge is a controlled evaluation of Arabic speech to text transcription, as well as supervised word alignment using Aljazeera TV channel recordings. This year MGB-2 uses a multi-dialect dataset, spanning more than 10 years of Arabic language broadcasts. The total amount of speech data crawled from Aljazeera using the QCRI Advanced Transcription System (QATS)\cite{ali2014qcri} was about 3,000 hours of broadcast programmes, whose durations ranged from 3--45 minutes. For the purpose of this evaluation, we  used only those programmes with transcription on their Arabic website, \url{Aljazeera.net}. These textual transcriptions contained no timing information. The quality of the transcription varied significantly: the most challenging were conversational programmes in which overlapping speech and dialectal usage was more frequent.

The Arabic MGB challenge had two main evaluation conditions:
\begin{itemize}[noitemsep]
\item Speech-to-text transcription of broadcast audio.
\item Alignment of broadcast audio to a provided transcription.
\end{itemize}
In this paper we describe the challenge data and provided metadata, with a focus on metadata refinement and data preparation. We discuss the two evaluation conditions in greater detail, and also outline the baseline systems provided for each task. We then outline the different systems that participants developed for the challenge, and give an overview for each of the provided system.

\section{MGB-2 Challenge Data}
\label{sec:format}

The Arabic MGB-2 Challenge used more than 1,200 hours of broadcast videos recorded during 2005--2015 from the Aljazeera Arabic TV channel. These programmes were manually transcribed, but not in a verbatim fashion. In some cases, the transcript includes re-phrasing, the removal of repetition, or summarization of what was spoken, in cases such as overlapping speech.  We found that the quality of the transcription varies significantly. The  WER between the original transcribed text from Aljazeera to the verbatim version is about 5\% on the development set. 

\subsection{Metadata challenges}
We have selected Aljazeera programmes that were manually transcribed (albeit without timing information). A total of 19 programmes series were collected, recorded over 10 years. Most, but not all, of the recorded programmes included the following metadata: programme name, episode title, presenter name, guests' names, speaker change information, date, and topic. 
The duration of an episode is typically 20--50 minutes, and the recorded programmes can be split into three broad categories; \textbf{conversation} (63\%), where a presenter talks with more than one guest discussing current affairs; \textbf{interview} (19\%), where a presenter speaks with one guest;  and \textbf{report} (18\%), such as news or documentary. Conversational speech, which includes the use of multiple dialects and overlapping talkers, is  a challenging condition and is the typical scenario for political debate and talk show programmes.

Much of the recorded data used in MGB-2 was Modern Standard Arabic (MSA): we estimate that more than 70\% of the speech is MSA, with the rest in Dialectal Arabic (DA), which we categorised as: Egyptian (EGY), Gulf (GLF), Levantine (LEV), and North African (NOR).  English and French language speech is also included, where typically the speech is translated and dubbed into Arabic.  This is not marked in the transcribed text.

\begin{figure}[ht]
\centering
\includegraphics [scale=0.55, frame]{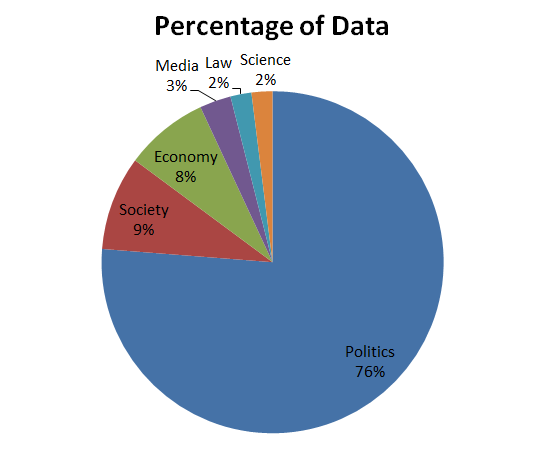}
\caption{Coarse-grained Domain Distribution}
\label{DomainFig}
\end{figure}

The original transcription has no clear metadata structure that would enable domain classification, so we decided to perform classification based on the keyword tags that are provided for the 3,000 episodes to define 12 domain classes, namely: politics, economy, society,  culture, media, law, science, religion, education, sport, medicine, and military. Because some domains have a very small number of programmes, we merged them to the nearest domain to have a coarse-grained classification as shown in Figure~\ref{DomainFig}, where the politics domain is the most frequent class.

\subsection{Data Processing and Light Alignment}
Removing programmes with damaged aligned transcriptions resulted in a total of about 1,200 hours of audio, which was released to MGB-2 participants.  All programmes were aligned using the QCRI Arabic LVCSR system~\cite{ali2014complete}, which is grapheme-based with one unique grapheme sequence per word .  It used LSTM acoustic models and trigram language models with a vocabulary size of about one million words. The same language model and decoding setup was used for all programmes.  For each programme, the ASR system generated word-level timings with confidence scores for each word. This ASR output was aligned with the original transcription to generate small speech segments of duration 5--30 seconds suitable for building speech recognition systems.

\begin{figure}[ht]
\centering
\includegraphics [scale=0.8, frame]{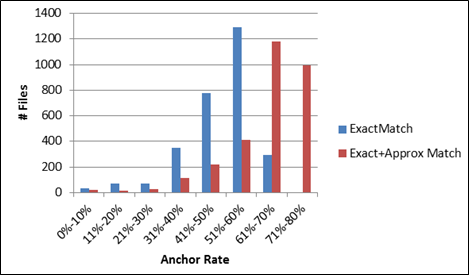}
\caption{Anchor Rates Distribution for Files}
\label{AnchorRateFiles}
\end{figure}

As shown in ~\cite{braunschweiler2010lightly} and based on the Smith\textendash Waterman algorithm~\cite{smith1981identification}, we  used to identify matching sequences by performing local sequence alignment to determine similar regions between two strings.  We addressed two challenges when aligning the data:
\begin{enumerate}[noitemsep]
\item The original transcription did not match the audio in some cases owing to edits to enhance clarity, paraphrasing, the removal of hesitations and disfluencies, and summarisation in cases such as overlapping speech.
\item Poor ASR quality in cases such as noisy acoustic environments, dialectal speech, use of out of vocabulary words, and overlapped speech.
\end{enumerate}
We  applied two levels of matching to deal with these challenges: exact match (where both transcription and ASR output are identical), and approximate match (where there is a forgiving edit distance between words in the transcription and ASR output).\\
To evaluate the quality of alignment between ASR output and the transcription, we calculated the ``anchor rate'' for each segment as follows:
\begin{equation}
Anchor Rate = \frac{\#MatchedWords}{\#TranscriptionWords}
\end{equation}
The AnchorRate across all segments came with the following; 48\% exact match, 15\% approximate match, and 37\% with no match. More details about the AnchorRate distribution is shown in figure~\ref{AnchorRateFiles}.



To assign time for non-matching word sequences, we used linear interpolation to force-align the original text to the remaining speech segments.

After aligning the whole transcript, each audio file was acoustically segmented into speech utterances, with a minimum silence duration of 300 milliseconds. The  metadata for aligned segments includes timing information obtained from the ASR, speaker name, and text  obtained from the manual transcription. For each segment, the Average Word Duration in seconds (AWD), the Phoneme Matching Error Rate (PMER), and Word Matching Error Rate (WMER) are stored in the given meta-data.

\begin{figure*}[ht]
\centering
\includegraphics [scale=0.5, frame]{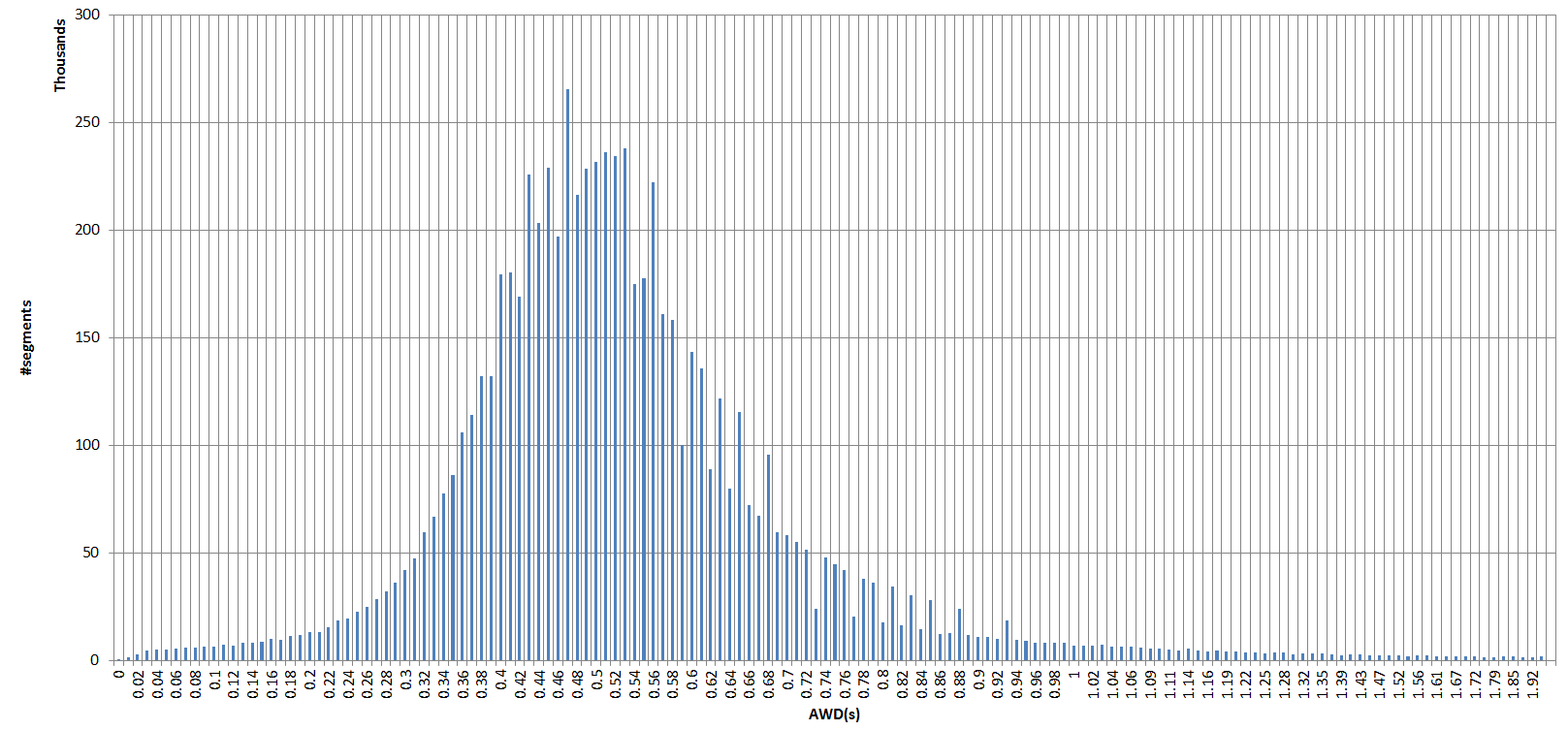}
\caption{Distribution of Average Word Duration (AWD)}
\label{AWD1Fig}
\end{figure*}

\begin{figure*}[ht]
\centering
\includegraphics [scale=0.5, frame]{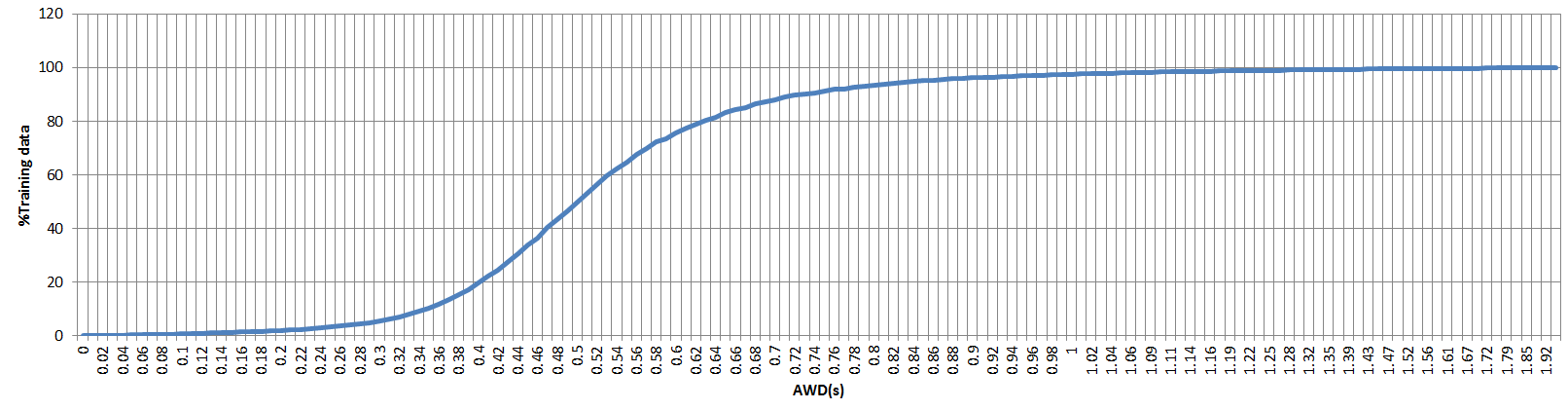}
\caption{Cumulative Average Word Duration} 
\label{AWD2Fig}
\end{figure*}

\begin{figure*}[ht]
\centering
\includegraphics [scale=0.7, frame]{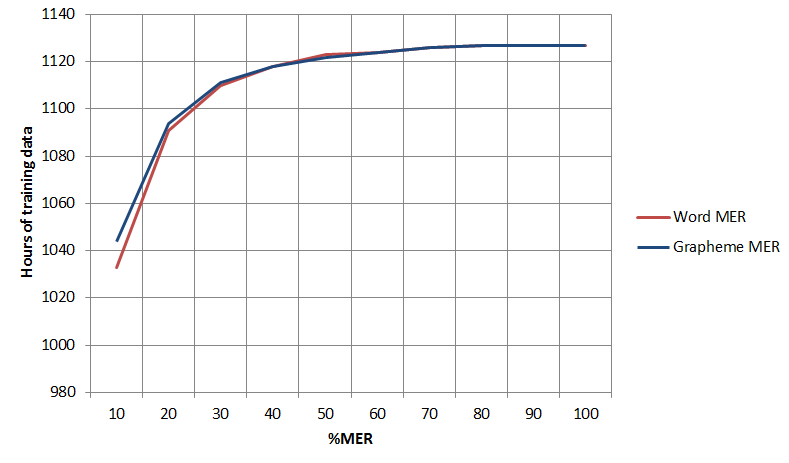}
\caption{Cumulative Duration for Word (red line), and Grapheme (blue line) Matching Error Rate (MER)}
\label{MERFig}
\end{figure*}

Overall, more than 550,000 segments with a total duration of 1,200 hours were made available together with the aligned transcription and metadata. Figure ~\ref{AWD1Fig} shows segment distribution according to AWD value, and Figure ~\ref{AWD2Fig} shows segment distribution according to cumulative AWD value. Figure ~\ref{MERFig} shows Cumulative duration for grapheme and word matching error rate.

During data preparation, we removed about 300 hours, mainly coming from very short audio clips with the corresponding full transcription. These audio clips are just the highlights of other programmes. No further filtering was applied to the 1,200 hours data.

\subsection{Lexicon}
Two lexicons were made available for participants in the challenge: A grapheme-based lexicon\footnote{http://alt.qcri.org/resources/speech/dictionary/ar\-ar\_grapheme\_lexicon\_2016\-02\-09.bz2} and with more than 900K entries with one unique grapheme sequence per word, and phoneme lexicon\footnote{http://alt.qcri.org/resources/speech/dictionary/ar\-ar\_lexicon\_2014\-03\-17.txt.bz2} with more than 500K words with an average of four phoneme sequence per word using our previous Vowelization to Phonetization (V2P) pipeline~\cite{ali2014complete}. Participants could also choose any lexicon outside the provided resources.

\subsection{Further Data Improvement}
We found that in some cases, the silence information coming from the ASR system was inaccurate and could lead to poor segmentation. This was mainly because the silence model in the basline ASR system~\cite{ali2014qcri}  acted as a garbage model, absorbing non-speech noise and sometimes overlapping speech.  A number of approaches could potentially improve the segmentation: a) run an ASR system with separate models for silence, and non-speech noise; b) employ an externally-trained Voice Activation Detection (VAD) system and apply on silences; c) train and use a seprate model for overlap speech detection; d) apply a word alignment algorithm to get better timing for each word using the text of each programme to build an in-domain LM.  We would not expect a better segmentation to make big impact on the quality of the ASR training data. However, it will be crucial for future challenges in dialect detection and potential dialectal diarization.

\section{Evaluation Tasks}
\label{sec:pagestyle}

The MGB-2 Challenge featured two evaluation tasks, transcription and alignment. For each the only allowable acoustic and language model training data was that specified above. To enable comparability, there was no option for participants to bring additional training data to the evaluation. Use of the provided other resources (e.g. dictionary) was optional.

 \subsection{Speech-to-text transcription}
 \label{subsection:speech-to-text}
 This is a standard speech transcription task operating on a collection of whole TV shows drawn from diverse Arabic dialectal programmes from Aljazeera TV channel. Scoring required ASR output with word-level timings. Segments with overlapped speech were scored but not considered in the main ranking. Overlapped speech was defined to minimise the regions removed -- at the segment level where possible. As the training data comes from only 19 series, some  programmes from the same series appeared in training, development and evaluation data.  
Each show in the development and evaluation set was  processed independently, so no speaker linking across shows was given. The data were carefully selected to cover different genres, and being diverse among the five dialects. Development and evaluation came from the last month of 2015 to avoid being seen in the training data.
The duration of each file was between 20 minutes and 50 minutes with a total duration of 10 hours for each set. These files were verbatim transcribed, and manually segmented for speech/silence/overlapped-speech with segment length between three and ten seconds. Words with hesitation or correction were also marked by adding a special symbol at the end of these words. \newline
Four WER numbers were reported for the speech-to-text task:
\begin{itemize}[noitemsep]
\item Scoring the original text as being produced by manual transcription, which might have punctuation/diacritization.
\item Scoring after removing any punctuation or diacritization.
\item Scoring using the Global Mapping File (GLM), which is the official results for the competition. This deals with various ways for writing numbers, and common words with no standard orthography.
\item Scoring after normalizing Alef, Yaa, and Taa Marbouta characters in the Arabic text, i.e. assuming that these kinds of differences are not considered as errors because they can be easily corrected using a surface spelling correction component.
\end{itemize}

\subsection{Alignment}
In this task, participants were supplied with a tokenised version of the transcription as published  on the broadcaster's website, without timing information. The task was to align the given transcription to the spoken audio at word level. The original transcription often differs from the actual spoken words. Scoring was performed by using a precision/recall measure, derived from the automatic alignment of a careful manual transcription. A word is considered to be a match if both start and end times fall within a 100 milliseconds window of the associated reference word. More details about alignment scoring can be found here ~\cite{bell2015mgb}.

\section{Baseline System}
\label{sec:typestyle}
We provided an open source baseline system for the challenge, via a GitHub repository\footnote{https://github.com/Qatar-Computing-Research-Institute/ArabicASRChallenge2016}. Data was shared in XML format\footnote{http://xmlstar.sourceforge.net/}. The baseline system included data pre-processing, data selection,  acoustic modelling (AM), and language modelling (LM), as well as decoding. This allowed participants to focus on more advanced aspects of ASR and LM modelling. A Kaldi toolkit~\cite{povey2011kaldi} recipe was made available for the MGB2, and for language modelling the SRILM~\cite{stolcke2002srilm} toolkit was used.
The baseline system was trained on 250 hours sampled from the training data, comes from 500 episodes. This system uses a standard MFCC multi-pass decoding:
\begin{itemize}[noitemsep]
\item The first pass uses a GMM with 5,000 tied states, and 100K total Gaussians, trained on features transformed with FMLLR
\item The second pass is trained using a DNN with four hidden layers, and 1024 neurons per layer, sequenced trained with the MPE criterion
\item A tri-gram language model is trained on the normalised version of the sample data text (250 hours).
\end{itemize}
The baseline results were reported on 10 hours verbatim transcribed development set: 34\% (8.5 hours) for the non-overlap speech and 73\% (1.5 hours) for the overlap speech.

\section{Submitted Systems and Results}
\label{sec:majhead}

Eight teams submitted systems to the MGB-2 challenge in the speech-to-text task. There has been no submission for the alignment task. We have attempted to highlight key features of various systems below. Detailed system descriptions are available at http://mgb-challenge.org/.\newline

\textbf{Task 1: Speech-to-text transcription}
\begin{itemize}[noitemsep,topsep=0pt]
  \item QCRI~\cite{KhuranaMGB2}: They have used 1,200 hours for training. They applied data augmentation with speed factors of 0.9,1.0, and 1.1. This gives them three times the original speech utterances. The speed perturbed data is followed by volume perturbation. For AM, they used three LF-MMI trained models; TDNN, LSTM and BLSTM. The three AM combined using MBR. For LM, they used three LMs; tri-gram for the first pass decoding and interpolated four-gram and RNN with MaxEnt connections for LM rescoring. Their final WER was 14.7\%.
    \item LIUM~\cite{TomashenkoMGB2}: They have aligned the training data, and applied data selection. Their total realigned data is about 650 hours. They used both grapheme and phoneme modelling. For their phoneme system, they have used automatic vowelizer followed by pronunciation rules with an average 1.6 pronunciation per lexicon entry. Their main contribution comes from data selection, training four neural network for AM of different types (DNN and TDNN), with various acoustic features (PLP, BNF, GMMD), and two types of phonetization. The final system, obtained through a confusion network combination of the four developed systems. Their final results were 16.7\%, and 15.7\% as a late submission.
\item MIT~\cite{AlhanaiMGB2}: They have used the 1,200 hours for training, they used several neural network topologies; feed-forward, CNN, TDNN, LSTM, H-LSTM, and G-LSTM. The models capturing temporal context (LSTMs) out-performed all other models. A discriminatively trained five layer G-LSTM was the best performing acoustic modelling. Their best results came from system combination of the top two hypothesis from the sequence trained G-LSTM models. Their final results were 18.3\%.
  \item NDSC~\cite{XuKuiMGB2}: They have selected data with zero WMER and an AWD ranging from 0.3s to 0.7s, and applied LTMD for further filtering. This gives them about 680 hours for training. For AM, they trained hyprid DNN, LSTM, and TDNN systems. Speed perturbation was used in the LSTM system. RNNLM was used for LM rescoring. They used MBR for their system combination which gives them an additional 0.9\% absolute reduction in WER. They have also investigate automatic segmentation based on long-term information, the final results using automatic segmentation has 1.2\% absolute increase in WER. Their final results were 19.4\% using the automatic segmentation, and 18.2\% using the manual segmentation for the official submission.\newline
\end{itemize}
The results for all the submitted systems are listed in Table~\ref{tab:res1}, and \ref{tab:res2}. For each submission we have four results:
WER1: scored with original text, WER2: scored after removing punctuation and vowelization, \textbf{WER3}: scored using the GLM, and WER4: scored after normalization. More details about the scoring is mentioned in section~\ref{subsection:speech-to-text}. \textbf{WER3} using the Global Mapping File is the official score.\newline

\textbf{Task 2: Alignment}
\begin{itemize}[noitemsep,topsep=0pt]
  \item[]There has not been any submission for the alignment task in the MGB-2, so, we share here the base line system for alignment with a precision of 0.83 and recall 0.7 and F measure of 0.76.  This has been calculate using 100 msec as window for accepting the timing. Total words in reference 55K words, and hypothesis has 47K words and match count is 39K words. The evaluation looks for matches within 100 milliseconds of the start/end times of each word which was difficult.
\end{itemize}

\begin{table}
\centering
\resizebox{\columnwidth}{!}{%
\begin{tabular}{r|llll}
\multicolumn{1}{r}{}
& \multicolumn{1}{l}{WER1}
& \multicolumn{1}{l}{WER2}
& \multicolumn{1}{l}{WER3}
& \multicolumn{1}{l}{WER4} \\  \cline{2-5}
QCRI& 23.7& 17.6 & 17.3 & 16.8 \\
LIUM& 25.5& 19.6 & 19.2 & 18.9 \\
MIT& 26.2 & 20.2 & 19.9 & 19.4 \\
NDSC & 29.5 & 24 & 23.8 & 23.4 \\
NHK& 39.1 & 34.4 & 34.2 & 33.9 \\
Cairo Univ& 45.6 & 41.5 & 41.2 & 40.9 \\
Seville Univ& 57.0 & 53.3 & 53.2 & 52.9 \\
Eqra& 58.5 & 56.5 & 56.4 & 53.1 
\end{tabular}%
}
\caption{\textit{Including Overlap Speech Results.}}\label{tab:res1}
\end{table}

\begin{table}
\centering
\resizebox{\columnwidth}{!}{%
\begin{tabular}{r|llll}
\multicolumn{1}{r}{}
& \multicolumn{1}{l}{WER1}
& \multicolumn{1}{l}{WER2}
& \multicolumn{1}{l}{\textbf{WER3}}
& \multicolumn{1}{l}{WER4} \\  \cline{2-5}
QCRI& 21.1& 15.0& \textbf{14.7}& 14.1 \\
LIUM&23.0& 17.0& \textbf{16.7}& 16.4 \\
MIT& 23.7& 17.5& \textbf{17.3}& 16.8 \\
NDSC& 24.4& 18.5& \textbf{18.2}& 17.8 \\
NHK& 34.7& 29.6& \textbf{29.5}& 29.1 \\
Cairo Univ& 43.3& 39.1& \textbf{38.8}& 38.6 \\
Seville Univ& 55.0& 51.2& \textbf{51.1}& 50.8 \\
Eqra& 56.8& 54.8& \textbf{54.7}& 51.3
\end{tabular}%
}
\caption{\textit{\textbf{Official} Results Excluding Overlap Speech.}}\label{tab:res2}
\end{table}
\section{Conclusions and Future Challenges}
The MGB-2 challenge continued the effort for using fixed training sets for acoustic and language modelling training for building speech recognition systems for broadcast media. This year's challenge focused on multi-dialectal Arabic. More than 1,200 hours with lightly supervised transcription have been shared coming from Aljazeera TV programs, along with more than 110 million words from \url{Aljazeera.net} web archive. 13 teams participated, with eight submitting a final system. On the speech-to-text task, the baseline WER was 34\%, and the best system submitted was 14.7\%. The top ranked systems used a combination of several deep and sequential neural network modelling for acoustic modelling. For language modelling, the main focus was on combining RNN with n-gram models. Both grapheme and phoneme units were investigated. The second task is was word alignment, for which we received no submissions; however, we developed a baseline system with precision of 0.83 and recall 0.7, which we hope will encourage submissions in future.  The authors are planning to continue the Arabic multi-dialect challenge, using the current tasks and training data, and develop it further into dialectal speech processing in terms of both dialect detection and dialectal speech recognition.
\label{sec:print}


\label{sec:ref}

\bibliographystyle{IEEEbib}
\bibliography{main}

\begin{thebibliography}{10}

\bibitem{bell2015mgb}
P.~J. Bell, M.~J.~F. Gales, T.~Hain, J.~Kilgour, P.~Lanchantin, X.~Liu,
  A.~McParland, S.~Renals, O.~Saz, M.~Wester, and P.~C. Woodland,
\newblock ``The {MGB} challenge: Evaluating multi-genre broadcast media
  recognition,''
\newblock in {\em ASRU}, 2015.

\bibitem{ali2014qcri}
A.~Ali, Y.~Zhang, and S.~Vogel,
\newblock ``{QCRI} advanced transcription system ({QATS}),''
\newblock in {\em SLT}, 2014.

\bibitem{ali2014complete}
A.~Ali, Y.~Zhang, P.~Cardinal, N.~Dahak, S.~Vogel, and J.~Glass,
\newblock ``A complete {K}aldi recipe for building {A}rabic speech recognition
  systems,''
\newblock in {\em SLT}, 2014.

\bibitem{braunschweiler2010lightly}
N.~Braunschweiler, M.~J.~F. Gales, and S~Buchholz,
\newblock ``Lightly supervised recognition for automatic alignment of large
  coherent speech recordings.,''
\newblock in {\em Interspeech}, 2010.

\bibitem{smith1981identification}
T.~Smith and M.~Waterman,
\newblock ``Identification of common molecular subsequences,''
\newblock {\em Journal of molecular biology}, vol. 147, no. 1, 1981.

\bibitem{povey2011kaldi}
D.~Povey, A.~Ghoshal, G.~Boulianne, L.~Burget, O.~Glembek, N.~Goel,
  M.~Hannemann, P.~Motlicek, Y.~Qian, P.~Schwarz, J.~Silovsky\`, G.~Stemmer,
  and K.~Vesely,
\newblock ``The {K}aldi speech recognition toolkit,''
\newblock in {\em ASRU}, 2011.

\bibitem{stolcke2002srilm}
Andreas Stolcke et~al.,
\newblock ``{SRILM} - an extensible language modeling toolkit.,''
\newblock in {\em Interspeech}, 2002.

\bibitem{KhuranaMGB2}
S.~Khurana and A.~Ali,
\newblock ``{QCRI} advanced transcription system ({QATS}) for the arabic
  multi-dialect broadcast media recognition: {MGB}-2 challenge,''
\newblock in {\em SLT}, 2016.

\bibitem{TomashenkoMGB2}
N.~Tomashenko, K.~Vythelingum, A.~Rousseau, and Y.~Est\`eve,
\newblock ``{LIUM} {ASR} systems for the 2016 multi-genre broadcast {A}rabic
  challenge,''
\newblock in {\em SLT}, 2016.

\bibitem{AlhanaiMGB2}
T.~AlHanai, W.~Hsu, and J.~Glass,
\newblock ``Development of the {MIT} {ASR} system for the 2016 {A}rabic
  multi-genre broadcast challenge,''
\newblock in {\em SLT}, 2016.

\bibitem{XuKuiMGB2}
X.~Yang, D.~Qu, W.~Zhang, and W.~Zhang,
\newblock ``The {NDSC} transcription system for the 2016 multi-genre broadcast
  challenge,''
\newblock in {\em SLT}, 2016.

\end{thebibliography}

\end{document}